\documentclass[runningheads]{llncs}

\usepackage[final,arxiv]{eccv}

\usepackage{eccvabbrv}

\usepackage{graphicx}
\usepackage{booktabs}
\usepackage{xcolor}

\usepackage{amsmath}
\usepackage{graphicx}
\usepackage{algorithm}
\usepackage{algpseudocode}

\usepackage{makecell}

\usepackage{multirow}
\usepackage[table]{xcolor} 

\usepackage{lipsum}
\usepackage{balance}
\usepackage{scalerel}
\usepackage{bm}

\usepackage{pifont}
\newcommand{\cmark}{{\color{Green}\ding{51}}}%
\newcommand{\xmark}{{\color{Red}\ding{55}}}%

\usepackage[accsupp]{axessibility}  

\usepackage{hyperref}

\usepackage{orcidlink}

\begin{document}

\title{Non-Minimal Sampling and Consensus\\for Prohibitively Large Datasets} 

\titlerunning{Non-Minimial Sampling and Consensus for Prohibitively Large Datasets}

\author{Seong Hun Lee\inst{1}\orcidlink{0000-0002-3821-1968} \and Patrick Vandewalle\inst{2}\orcidlink{0000-0002-7106-8024} \and
Javier Civera\inst{1}\orcidlink{0000-0003-1368-1151}}

\authorrunning{S.~H.~Lee et al.}

\institute{
I3A, University of Zaragoza, 50018 Zaragoza, Spain
\email{\{seonghunlee,jcivera\}@unizar.es}
\and 
EAVISE, KU Leuven, 2860 Sint-Katelijne-Waver, Belgium
\email{patrick.vandewalle@kuleuven.be}}

\maketitle

\begin{abstract}
We introduce NONSAC (Non-Minimal Sampling and Consensus), a general framework for robust and scalable model estimation from arbitrarily large datasets contaminated with noise and outliers.
NONSAC repeatedly samples non-minimal subsets of data and generates model hypotheses using a robust estimator, producing multiple candidate models.
The final model is selected based on a predefined scoring rule that evaluates hypothesis quality.
Our framework is estimator-agnostic and can be integrated with existing geometric fitting algorithms such as RANSAC to improve both scalability and robustness to outliers.
We propose and evaluate various scoring rules for NONSAC on relative camera pose estimation, Perspective-n-Point, and point cloud registration.
Furthermore, we showcase the applicability of NONSAC to correspondence-free point cloud registration by hypothesizing all-to-all correspondences.
\end{abstract}

\section{Introduction}
\label{sec:intro}
\begin{figure*}[ht!]
 \centering
 \includegraphics[width=\textwidth]{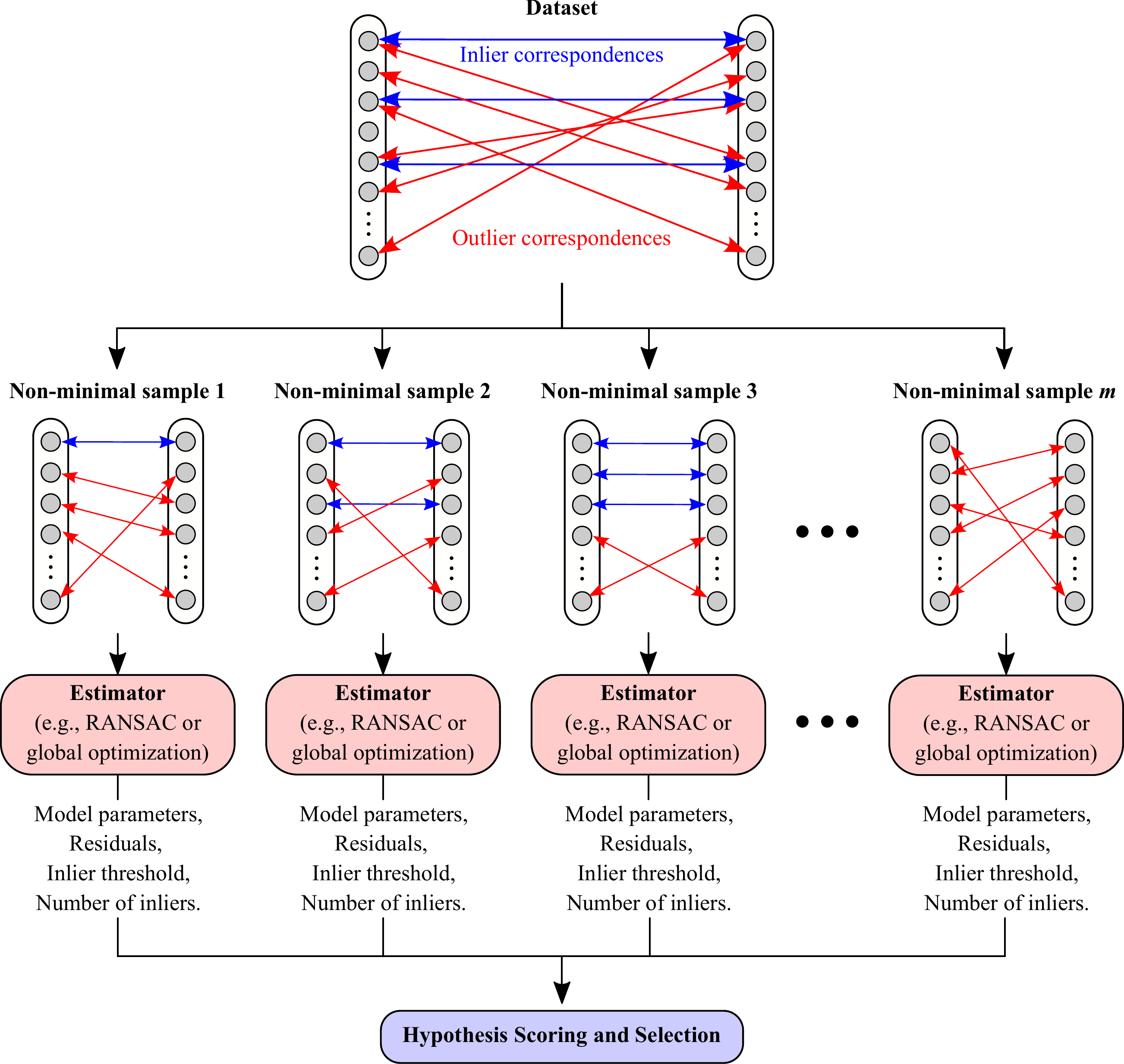}
\caption{NONSAC pipeline: 
From the original correspondence dataset, we randomly draw $m$ non-minimal samples. 
Due to randomness, some samples are likely to contain more inliers than others. 
In this example, Sample 3 has the highest number of inliers. 
For each sample, we apply an off-the-shelf estimator to compute the model parameters, residuals, inlier counts, and related values. 
We then aggregate the results from all samples to select the most reliable hypothesis according to a chosen scoring rule. 
In this case, the hypothesis obtained from Sample 3 is expected to be among the most accurate and thus to receive the highest score and be selected.
}
\label{fig:pipeline}
\end{figure*}

Robust geometric model estimation is a cornerstone of computer vision and robotics, underlying tasks such as relative camera pose estimation, Perspective-n-Point (PnP), and point cloud registration. 
In practice, these problems are typically formulated as estimating model parameters from a large set of putative correspondences contaminated with noise and a significant fraction of outliers. 
Classical robust estimators, most notably RANSAC \cite{fischler1981random} and its variants \cite{loransac, ansac,usac,gcransac,magsac_plus}, have proven highly effective when datasets are of moderate size and the outlier ratio is manageable. 
However, some modern applications produce extremely large correspondence sets, for example through dense matching \cite{sift_flow, raft, loftr} or by hypothesizing all-to-all correspondences \cite{fredriksson_2016_cvpr,enqvist_2009_iccv,teaser}, which can lead to prohibitive memory requirements and extreme outlier ratios. 
In such cases, many existing correspondence-based methods either become computationally infeasible or suffer substantial performance degradation. 
This motivates the need for a scalable and robust framework that can operate reliably under both massive data volumes and severe outlier contamination.

In this work, we propose NONSAC (\textbf{Non}-minimal \textbf{Sa}mpling and \textbf{C}onsensus), a general framework for correspondence-based geometric model estimation that can be integrated with existing estimators to enhance their scalability and robustness. 
Fig.~\ref{fig:pipeline} shows the pipeline of NONSAC.
Our main contributions are three-fold:
\begin{enumerate}
    \item We illustrate the necessity and advantages of non-minimal sampling when dealing with prohibitively large datasets that contain a high proportion of outliers. 
    With a robust core estimator and a fixed number of non-minimal samples, NONSAC can process arbitrarily large datasets efficiently and robustly, without inherent scalability constraints.

    \item We propose and evaluate several rules for hypothesis scoring and selection (see Fig.~\ref{fig:pipeline}).
    We conduct extensive experiments to assess and compare these rules across three fundamental problems in computer vision and robotics: relative pose estimation (with 2D-to-2D correspondences), Perspective-n-Point (with 3D-to-2D correspondences), and point cloud registration (with 3D-to-3D correspondences). 

    \item We further demonstrate the effectiveness of NONSAC on the correspondence-free point cloud registration problem using a real dataset reconstructed from 3D scanning \cite{bunny}. 
    In this setting, we hypothesize all-to-all correspondences between two unmatched point clouds, resulting in an extremely large set of candidate matches and a very high outlier ratio, specifically $2.5\times10^5$ correspondences with 99.9\% outliers in our experiment. 
    To the best of our knowledge, no prior method has demonstrated the ability to handle data of this scale under such extreme levels of outlier contamination within a reasonable runtime.
    
\end{enumerate}

\section{Related Work}

\noindent$\bullet$ \textbf{General-purpose robust estimation techniques}:
NONSAC is a general framework that can be applied to a wide range of problems.
A qualitative comparison between NONSAC and existing frameworks is provided in Tab.~\ref{tab:method_comparison}.
In contrast to RANSAC-based methods (\eg, \cite{fischler1981random, loransac, ansac, usac}) and global optimization methods (\eg, \cite{irls, gnc, bnb, aoxiang_2022_tpami}), NONSAC never evaluates the estimated model on the entire dataset, which makes it scalable to arbitrarily large datasets.
LO-RANSAC~\cite{loransac} and ANSAC~\cite{ansac} are closely related to NONSAC, as they also use non-minimal samples.
LO-RANSAC improves RANSAC by refining promising models using non-minimal samples drawn from their inlier sets, while ANSAC improves the hypothesis generation stage by adaptively drawing non-minimal samples from ranked correspondences.
NONSAC, in contrast, draws fixed-size non-minimal samples directly from the original dataset, processes them independently, and evaluates them using robust scoring.
RHT \cite{rht} and hypothesis averaging methods (\eg, via rotation averaging \cite{hartley_l1, lee_rsa_revisited}) are scalable to datasets of arbitrary size.
However, these approaches generate hypotheses from minimal samples and evaluate them relative to other hypotheses, which generally makes them less robust than NONSAC, where hypotheses are both generated and evaluated using non-minimal samples.

\noindent$\bullet$ \textbf{Problem-tailored geometric model estimation}:
Relative pose estimation, PnP, and point cloud registration are fundamental problems in 3D vision with extensive literature tailored to each of them.
Recent advances in relative pose estimation include QCQP/SDP-based methods \cite{briales_2018_cpvr,zhao,garcia_2021_jmiv,tirado_2024_CVPR}, learning-based approaches \cite{yi_2018_cvpr, moran2024consensus}, and other optimization-based methods \cite{Cai2024LinearRP, aoxiang_2022_tpami}.
For the PnP problem, recent works span inlier-only optimization methods \cite{sqpnp,DEPnP,acepnp}, learning-based approaches \cite{campbell_2020_eccv,chen_2025_tpami}, and robust estimation methods \cite{long_ral_2025, xu_2025_tip}.
For point cloud registration, recent methods include consistency graph-based approaches \cite{teaser, sc2pcr,mac, pcr99}, learning-based methods \cite{pointdsc, vbreg}, and other robust optimization techniques \cite{trde, tear}.

We emphasize that neither the general-purpose frameworks nor the problem-specific methods discussed in this section should be viewed as directly competing with NONSAC. 
Instead, any of these methods can be incorporated into NONSAC as a core estimator, as demonstrated in Sec.~\ref{sec:results}, where we integrate RANSAC \cite{fischler1981random} and PCR-99 \cite{pcr99} into the NONSAC framework.

\begin{table*}[t]
\resizebox{\textwidth}{!}{
\begin{tabular}{ccccc}
\hline
 & Random  & Non-minimal & Scales to & Works well in \\
 & sampling? & batch & arbitrarily  & high dimensional \\
  &   & estimation? & large datasets? & parameter spaces? \\
\hline
\multicolumn{1}{r}{\rule{0pt}{3ex}RANSAC \cite{fischler1981random}} & \cmark & \xmark  & \xmark &\cmark\\
\multicolumn{1}{r}{\rule{0pt}{3ex}LO-RANSAC \cite{loransac}, ANSAC \cite{ansac}} & \cmark & \cmark  & \xmark &\cmark\\
\multicolumn{1}{r}{\makecell[{{r}}]{\rule{0pt}{3ex}Robust global methods \\ (\eg IRLS \cite{irls}, GNC \cite{gnc}, BnB \cite{bnb})}} & \xmark & \cmark  & \xmark &\cmark\\
\multicolumn{1}{r}{\rule{0pt}{3ex}Random Hough Transform (RHT) \cite{rht}} & \cmark & \xmark  & \cmark &\xmark\\
\multicolumn{1}{r}{\rule{0pt}{3ex}Hypothesis averaging (\eg \cite{hartley_l1, lee_rsa_revisited})} & \cmark & \xmark  & \cmark &\cmark\\
\multicolumn{1}{r}{\rule{0pt}{3ex}NONSAC (Ours)} & \cmark & \cmark  & \cmark &\cmark\\
\hline
\end{tabular}
}
\caption{
Qualitative comparison of robust geometric model fitting methods.
}
\label{tab:method_comparison}
\end{table*}

\section{Method}
\label{sec:method}
\subsection{Non-minimal sampling}
\label{subsec:sampling}
As illustrated in Fig.~\ref{fig:pipeline}, NONSAC consists of three main steps: (1) non-minimal random sampling, (2) model estimation for each sample, and (3) hypothesis scoring and selection.
Non-minimal sampling of the original data has two main practical advantages:
\begin{enumerate}
    \item It enables us to process extremely large datasets incrementally rather than all at once, which would otherwise be prohibitively expensive in terms of computation time and memory consumption.
    By keeping the size of the non-minimal sample manageable, we avoid taking on more than we can reasonably handle.
    
    \item Some non-minimal samples may exhibit higher inlier ratios than the full original dataset, potentially enabling more accurate model estimation than when processing everything altogether.
    To be specific, around half of the non-minimal samples would exhibit higher inlier ratios than the full dataset.
    This is because, according to the Central Limit Theorem \cite{devore_statistics}, the sampling distribution of the sample proportion is approximately normal\footnote{The necessary condition for this approximation is that both $NP$ and $N(1-P)$ are greater than equal to 10 \cite{devore_statistics}.} with mean $P$ and variance $P(1-P)/N$, where $P$ is the proportion of the population (in our case, the inlier ratio) and $N$ is the sample size.
\end{enumerate}

\subsection{Scoring rules}
\label{subsec:scoring}
We consider the following scoring rules for model selection:
\begin{enumerate}
    \item \textit{Ideal hypothesis selection}: Choose the model closest to the ground truth. 
    This provides the upper bound achievable with non-minimal sampling.

    \item \textit{Fixed non-minimal sample}: Rather than drawing new non-minimal samples in each NONSAC iteration, we reuse the initially selected non-minimal sample throughout all iterations.\footnote{For example, suppose that we adopt RANSAC \cite{fischler1981random} as the core estimator, with 100 minimal samples drawn for each RANSAC run and 10 non-minimal samples of 1000 points used for NONSAC.
    In this case, the \textit{fixed non-minimal sample} rule essentially means that we take only one non-minimal sample of 1000 points and draw $10\times100=1000$ minimal samples from this 1000-point set.}
    
    \item \textit{Most inliers}: Choose the model where the largest number of inliers are found by the estimator.
    \item \textit{Closest pair}: Find the closest pair of models, \ie, model $i$ and $j$ where the distance between them, $d_
    {ij}$, is the smallest among all possible pairs.
    Between the two models, choose the one with more inliers.
    \item \textit{Closest triplet}: Find the closest triplet of models, \ie, model $i$, $j$ and $k$ where $\max(d_
    {ij}, d_{ik}, d_{jk})$ is the smallest among all possible triplets. Between the three models, choose the one with the most inliers.
    \item \textit{Minimum mean, median or Q3}: Choose the model with the smallest mean, median or upper quartile of inlier residuals.
    \item \textit{Pair cost}: Find the pair of models ($i$, $j$) that maximizes the following cost function:
    \begin{equation}
        c_{ij} = \frac{\max(n_i,n_j)}{\left(d_{ij}\right)^k},
    \end{equation}
    where $n_i$ and $n_j$ are the number of inliers, $d_{ij}$ is the distance between the two models, and $k$ is a parameter we set.
    After identifying the pair $(i, j)$ with the minimum cost value, select the model (either $i$ or $j$) that has the greater number of inliers.
    \item \textit{TLP cost}: Choose the model with the smallest truncated $L_p$ cost, \ie,
    \begin{equation}
    c_i = \sum_{j\in\mathcal{P}} \min\left(|r_j|^p, \tau^p\right),
    \end{equation}
    where $\mathcal{P}$ is the set of all point correspondences, $r_j$ is the residual of the $j$-th correspondence (in model $i$), $\tau$ is the truncation threshold, and $p$ is a parameter we set.
\end{enumerate}

\section{Results}
\label{sec:results}
We evaluate the performance of NONSAC with the scoring rules described in Section \ref{subsec:scoring}.
We apply our method to three fundamental problems in computer vision and robotics: relative camera pose estimation, Perspective-n-Point (PnP), and point cloud registration.
In all three problems, we use the rotation difference (in degrees) for the distance function $d_{ij}$ between model $i$ and $j$.
As for the error metric, we use the mean average accuracy (mAA) \cite{yi_2018_cvpr, jin_2021_ijcv} in the following form:
\begin{equation}
\label{eq:mAA}
    \textrm{mAA} = \frac{1}{10N_s}\sum_{\theta=1^\circ}^{10^\circ}\sum_{i=1}^{N_{\scaleto{s}{3pt}}}f(\theta, e_r),
\end{equation}
where $N_s$ is the number of Monte Carlo simulations ($100$ in all our experiments), $e_r$ is the rotation error (in degrees) with respect to the ground truth, and
\begin{equation}
    f(\theta, e_r) = 
    \begin{cases}
    1, & \text{if } e_r < \theta \\
    0, & \text{otherwise}\\
  \end{cases}.
\end{equation}

\subsection{Relative pose estimation}
\label{subsec:relative}
In this section, we apply NONSAC to the relative pose estimation problem.
To study the effect of noise level, outlier ratio, and sample size, we set up the simulations as follows:
We fix one camera at the origin and generate $n$ random points in front of the camera such that their normalized image points $(u_i, v_i)$ and depths $d_i$ follow uniform distributions in the range $[-1, 1]$ and $[0.1, 10]$, respectively.
We place the second camera at some position and orientation such that (1) the distance from the first camera is $1$ unit, (2) all of the $n$ points have depths greater than $0.1$ unit in the second camera, and (3) their normalized image points fall in the range $[-1, 1]$.
Noise with standard deviation $\sigma$ is added to the normalized image points in each camera, and some of them are replaced by uniformly distributed random outliers.

Next, we configure NONSAC as follows:
We pick $m$ random samples of $1000$ points.
From each sample, we estimate the relative pose between the cameras using a 5-point algorithm \cite{kukelova_2008_bmvc} in a standard RANSAC framework \cite{fischler1981random}.
We identify the inliers by applying a threshold of $(5\sigma)^2$ to the Sampson errors \cite{sampson_1982_cgip,luong_1996_ijcv,rydell_2024_cvpr}. 

Note that the probability of any two non-minimal samples having at least one point in common decreases with the total number of points, $n$.
To simulate extremely large datasets without actually generating an excessive amount of points, we set $n=1000m$ and simply ensure that the samples have zero overlap among themselves.
We evaluate our method by varying the image noise level ($\sigma$), the outlier ratio, the number of non-minimal samples ($m$), and the scoring rule from Sec. \ref{subsec:scoring}.

Table \ref{tab:relative} presents the results.
It shows that there is no clear winner that consistently yields the best results.
For large noise levels or low outlier ratios, the \textit{minimum mean} and \textit{minimum Q3} scoring rules perform best; however, they degrade significantly in small-noise, high-outlier-ratio scenarios.
In contrast, the \textit{pair cost ($k=0.05, 0.1$)} scoring rules perform best in these cases, but underperform at large noise levels.
Overall, the \textit{TLP cost ($p=0.1, 0.2$)} achieves a well-balanced accuracy across all tested scenarios, despite often falling slightly short of the \textit{minimum Q3}.
The averaged results in Tab.~\ref{tab:avg} further confirm the overall superiority of the \textit{TLP cost ($p=0.1, 0.2$)} scoring rules.

\begin{table*}[t]
\resizebox{\textwidth}{!}{
\begin{tabular}{l*{9}{>{\centering\arraybackslash}p{4em}}}
\toprule
& \multicolumn{3}{c}{\textbf{Relative pose estimation}}
& \multicolumn{3}{c}{\textbf{Perspective-n-Point}}
& \multicolumn{3}{c}{\textbf{Point cloud registration}} \\
& \multicolumn{3}{c}{Outlier ratio: 65\%--90\%}
& \multicolumn{3}{c}{Outlier ratio: 92\%--97\%}
& \multicolumn{3}{c}{Outlier ratio: 99\%--99.5\%}\\
& \multicolumn{3}{c}{\# of non-minimal samples:}
& \multicolumn{3}{c}{\# of non-minimal samples:}
& \multicolumn{3}{c}{\# of non-minimal samples:} \\
& 10  & 20  & 30 
& 10  & 20  & 30 
& 10  & 20  & 30  \\
\midrule

Ideal hypothesis selection          & \underline{0.91} & \underline{0.94} & \underline{0.96} & \underline{0.88} & \underline{0.94} & \underline{0.96} & \underline{0.82} & \underline{0.91} & \underline{0.95} \\
Fixed non-minimal sample    & 0.79 & 0.81 & 0.83 & \textcolor{blue}{\textbf{0.84}} & 0.88 & 0.92 & 0.47 & 0.54 & 0.60 \\
Most inliers    & 0.80 & 0.83 & 0.85 & 0.83 & 0.89 & 0.91 & 0.80 & 0.88 & 0.91 \\
Closest pair    & 0.75 & 0.75 & 0.78 & 0.78 & 0.86 & 0.87 & 0.72 & 0.83 & 0.89 \\
Closest triplet & 0.77 & 0.77 & 0.80 & 0.80 & 0.88 & 0.90 & 0.71 & 0.81 & 0.88 \\
Minimum mean        & \textcolor{blue}{\textbf{0.83}} & \textcolor{blue}{\textbf{0.87}} & 0.89 & 0.42 & 0.46 & 0.47 & 0.09 & 0.05 & 0.03 \\
Minimum median         & 0.81 & 0.86 & 0.88 & 0.40 & 0.42 & 0.41 & 0.09 & 0.04 & 0.03 \\
Minimum Q3          & 0.82 & \textcolor{blue}{\textbf{0.87}} & 0.89 & 0.56 & 0.61 & 0.61 & 0.12 & 0.06 & 0.05 \\
Pair cost ($k=0.02$)  & 0.81 & 0.84 & 0.86 & 0.83 & 0.90 & 0.91 & 0.80 & 0.88 & 0.91 \\
Pair cost ($k=0.05$)  & 0.82 & 0.86 & 0.87 & 0.83 & 0.90 & 0.91 & 0.80 & 0.88 & 0.91 \\
Pair cost ($k=0.1$)   & 0.82 & 0.86 & 0.87 & 0.83 & 0.90 & 0.92 & 0.80 & 0.88 & 0.91 \\
Pair cost ($k=0.2$)   & 0.80 & 0.82 & 0.84 & 0.83 & 0.90 & 0.92 & 0.80 & 0.88 & \textcolor{blue}{\textbf{0.92}} \\
Pair cost ($k=0.5$)   & 0.82 & 0.85 & 0.86 & 0.83 & 0.89 & 0.91 & 0.80 & 0.88 & \textcolor{blue}{\textbf{0.92}} \\
Pair cost ($k=1$)    & 0.78 & 0.80 & 0.81 & 0.82 & 0.88 & 0.90 & 0.78 & 0.86 & 0.91 \\
TLP cost ($p=0.01$)    & 0.78 & 0.80 & 0.83 & 0.56 & 0.64 & 0.65 & \textcolor{blue}{\textbf{0.81}} & \textcolor{blue}{\textbf{0.89}} & \textcolor{blue}{\textbf{0.92}} \\
TLP cost ($p=0.1$)         & \textcolor{blue}{\textbf{0.83}} & \textcolor{blue}{\textbf{0.87}} & \textcolor{blue}{\textbf{0.90}} & 0.83 & 0.90 & \textcolor{blue}{\textbf{0.93}} & \textcolor{blue}{\textbf{0.81}} & \textcolor{blue}{\textbf{0.89}} & \textcolor{blue}{\textbf{0.92}} \\
TLP cost ($p=0.2$)         & \textcolor{blue}{\textbf{0.83}} & \textcolor{blue}{\textbf{0.87}} & \textcolor{blue}{\textbf{0.90}} & 0.83 & 0.90 & \textcolor{blue}{\textbf{0.93}} & \textcolor{blue}{\textbf{0.81}} & \textcolor{blue}{\textbf{0.89}} & \textcolor{blue}{\textbf{0.92}} \\
TLP cost ($p=0.3$)         & \textcolor{blue}{\textbf{0.83}} & \textcolor{blue}{\textbf{0.87}} & 0.89 & 0.83 & 0.90 & \textcolor{blue}{\textbf{0.93}} & \textcolor{blue}{\textbf{0.81}} & \textcolor{blue}{\textbf{0.89}} & \textcolor{blue}{\textbf{0.92}} \\
TLP cost ($p=0.5$)         & \textcolor{blue}{\textbf{0.83}} & \textcolor{blue}{\textbf{0.87}} & 0.89 & 0.83 & \textcolor{blue}{\textbf{0.91}} & \textcolor{blue}{\textbf{0.93}} & \textcolor{blue}{\textbf{0.81}} & \textcolor{blue}{\textbf{0.89}} & \textcolor{blue}{\textbf{0.92}} \\
TLP cost ($p=1$)           & \textcolor{blue}{\textbf{0.83}} & \textcolor{blue}{\textbf{0.87}} & 0.89 & 0.83 & \textcolor{blue}{\textbf{0.91}} & \textcolor{blue}{\textbf{0.93}} & \textcolor{blue}{\textbf{0.81}} & 0.88 & \textcolor{blue}{\textbf{0.92}} \\
TLP cost ($p=2$)           & 0.82 & 0.86 & 0.88 & 0.83 & \textcolor{blue}{\textbf{0.91}} & \textcolor{blue}{\textbf{0.93}} & 0.80 & 0.88 & 0.91 \\
\bottomrule
\end{tabular}
}
\caption{Average mAA comparison of different scoring rules under varying numbers of non-minimal samples. Each entry in this table represents the average MAA computed across all evaluated outlier ratios and noise levels in Tabs. \ref{tab:relative}, \ref{tab:pnp}, and \ref{tab:pcr}.
\underline{Underline}: Ideal accuracy obtainable if one chooses the model closest to the ground truth. \textcolor{blue}{\textbf{Blue bold}}: Best results (excluding Ideal).}
\label{tab:avg}
\end{table*}

\subsection{Perspective-n-Point}
\label{subsec:pnp}
In this section, we apply NONSAC to the PnP problem.
To study the effect of noise level, outlier ratio, and sample size, we simulate a camera, 3D points and their normalized image points using the same procedure described in Sec. \ref{subsec:relative}.
We configure NONSAC in the same manner, except that we use P3P \cite{lee_p3p} within RANSAC and identify the inliers by applying a threshold of $5\sigma$ to the reprojection errors \cite{hartley_textbook}.
As in Sec. \ref{subsec:relative}, we simulate extremely large datasets by ensuring that the non-minimal samples have zero overlap among themselves.
The size of each non-minimal sample is set to 1000 points.

Table \ref{tab:pnp} presents the results.
As in the previous relative pose estimation results, the PnP results also show that there is no clear winner.
The \textit{minimum mean/median/Q3} scoring rules perform poorly most of the time, except in large-noise, small-outlier-ratio scenarios. 
The \textit{pair cost ($k=0.01$)} scoring rule produces good results when (i) the noise level is small ($\sigma=0.005$) or (ii) the noise level is moderate ($\sigma=0.01$) and the outlier ratio is high ($\geq95\%$). 
That said, the \textit{TLP cost} scoring rules achieve the best overall accuracy, with the suitable value for $p$ varying per scenario. 
We obtain the best results with $p=2$ when (i) the noise level is small ($\sigma=0.005$) or (ii) the noise level is moderate ($\sigma=0.01$) and the number of samples is small ($10$ in our experiment).
On the other hand, we get the best results with $p=0.5, 1$ when (i) the noise level is large ($\sigma=0.02$) or (ii) the noise level is moderate ($\sigma=0.01$) and the number of samples is large ($20$ and $30$ in our experiment).
The averaged results shown in Tab.~\ref{tab:avg} also suggest that the \textit{TLP cost ($p=0.5, 1, 2$)} scoring rules are relatively reliable when used with $20$ or more non-minimal samples.

\subsection{Point cloud registration}
\label{subsec:pcr}
In this section, we apply NONSAC to the point cloud registration problem.
To study the effect of noise level, outlier ratio, and sample size, we set up the simulations as follows:
First, we obtain the ground-truth point cloud by generating $n$ random 3D points, uniformly distributed inside a unit cube.
To obtain the query point cloud, we transform it using a random rigid body transformation, add Gaussian noise with standard deviation $\sigma$, and replace some of the points with random outlier points, uniformly distributed inside a 3D sphere circumscribing the unit cube.
As in Sec. \ref{subsec:relative} and \ref{subsec:pnp}, we simulate extremely large datasets by ensuring that the non-minimal samples have zero overlap among themselves.
The size of each non-minimal sample is set to 2000 points.
As for the estimator, we use a modified version of PCR-99 \cite{pcr99} with the following changes compared to the original version:
\begin{itemize}
    \item Instead of the log ratio \cite{pcr99}, we use the pairwise distance between 3D points to prescreen bad 3-point samples, and also, to score the correspondences.
    \item Instead of terminating the iterations based on the number of the number of inliers found, we do it when the number of valid 3-point samples passing the prescreening test (\ie, $n_\mathrm{hypothesis}$ in Alg. 1 of \cite{pcr99}) reaches 1000 or when the number of total 3-point samples drawn reaches 10,000.\footnote{As for the \textit{fixed non-minimal sample} rule, the number of total 3-point samples is changed to $10000\times\text{\# of non-minimal samples}$. We do not, however, change $n_\mathrm{hypothesis}$, because otherwise this can stall the algorithm when the number of 3-point samples that can pass the prescreening test is smaller than $1000\times\text{\# of non-minimal samples}$.}
\end{itemize}

Table \ref{tab:pcr} presents the results. 
When the outlier ratio is below $99.2\%$, we see that \textit{pair cost ($k=0.2, 0.5$)} scoring rules perform relatively well, often better than the \textit{TLP cost}.
In other cases, however, the \textit{TLP cost ($p= 0.01$--$0.5$)} scoring rules achieve higher accuracy. 
As in relative pose estimation and P$n$P, the highest overall accuracy for point cloud registration is achieved using the \textit{TLP cost}.
This is further confirmed by the averaged results reported in Tab.~\ref{tab:avg}.

\subsection{Correspondence-free point cloud registration}
\label{subsec:correspondence-free}

\begin{table}[t]
\resizebox{\textwidth}{!}{
\renewcommand{\arraystretch}{0.95}
\begin{tabular}{l*{4}{>{\centering\arraybackslash}p{5em}}}
\toprule
& \multicolumn{4}{c}{\textbf{Correspondence-free point cloud registration}} \\
\cmidrule(l){2-5}
& mAA($5^\circ$) & mAA($10^\circ$) & mAA($15^\circ$) & mAA($20^\circ$) \\
\midrule
\multicolumn{1}{c}{RANSAC \cite{fischler1981random}, GORE \cite{gore}, TR-DE \cite{trde}}
& \multicolumn{4}{c}{Excessive runtime ($>1$ hour)} \\[1ex]
\multicolumn{1}{c}{\textit{Consistency graph}-based methods}
& \multicolumn{4}{c}{\multirow{2}{*}{Out-of-memory}} \\
\multicolumn{1}{c}{\cite{sc2pcr, teaser, mac, pcr99}}
& \multicolumn{4}{c}{} \\[1ex]
\multicolumn{1}{c}{\textit{Deep learning}-based methods \cite{pointdsc, vbreg}}
& \multicolumn{4}{c}{Out-of-memory} \\[1ex]
\multicolumn{1}{c}{NONSAC with 10 non-minimal samples} \\
\hspace{5em} 1.\ Ideal hypothesis selection & \underline{0.75} & \underline{0.87} & \underline{0.92} & \underline{0.94} \\
\hspace{5em} 2.\ Most inliers               & 0.53 & 0.76 & 0.84 & 0.88 \\
\hspace{5em} 3.\ Closest pair               & 0.55 & 0.75 & 0.83 & 0.88 \\
\hspace{5em} 4.\ Closest triplet            & 0.58 & 0.79 & 0.86 & 0.89 \\
\hspace{5em} 5.\ Minimum mean               & 0.54 & 0.71 & 0.78 & 0.83 \\
\hspace{5em} 6.\ Minimum median             & 0.47 & 0.67 & 0.75 & 0.80 \\
\hspace{5em} 7.\ Minimum Q3                 & 0.38 & 0.60 & 0.70 & 0.76 \\
\hspace{5em} 8.\ Pair cost ($k=0.02$)       & 0.55 & 0.77 & 0.85 & 0.89 \\
\hspace{5em} 9.\ Pair cost ($k=0.05$)       & 0.56 & 0.78 & 0.85 & 0.89 \\
\hspace{5em} 10.\ Pair cost ($k=0.1$)       & 0.58 & 0.79 & 0.86 & 0.90 \\
\hspace{5em} 11.\ Pair cost ($k=0.2$)       & 0.59 & 0.79 & 0.86 & 0.90 \\
\hspace{5em} 12.\ Pair cost ($k=0.5$)       & 0.59 & 0.79 & 0.86 & 0.89 \\
\hspace{5em} 13.\ Pair cost ($k=1$)         & 0.58 & 0.77 & 0.85 & 0.89 \\
\hspace{5em} 14.\ TLP cost ($p=0.01$)       & \textcolor{blue}{\textbf{0.64}} & \textcolor{blue}{\textbf{0.82}} & \textcolor{blue}{\textbf{0.88}} & \textcolor{blue}{\textbf{0.91}} \\
\hspace{5em} 15.\ TLP cost ($p=0.1$)        & \textcolor{blue}{\textbf{0.64}} & \textcolor{blue}{\textbf{0.82}} & \textcolor{blue}{\textbf{0.88}} & \textcolor{blue}{\textbf{0.91}} \\
\hspace{5em} 16.\ TLP cost ($p=0.2$)        & 0.63 & \textcolor{blue}{\textbf{0.82}} & \textcolor{blue}{\textbf{0.88}} & \textcolor{blue}{\textbf{0.91}} \\
\hspace{5em} 17.\ TLP cost ($p=0.3$)        & 0.63 & 0.81 & 0.87 & \textcolor{blue}{\textbf{0.91}} \\
\hspace{5em} 18.\ TLP cost ($p=0.5$)        & 0.62 & 0.81 & 0.87 & 0.90 \\
\hspace{5em} 19.\ TLP cost ($p=1$)          & 0.61 & 0.80 & 0.87 & 0.90 \\
\hspace{5em} 20.\ TLP cost ($p=2$)          & 0.60 & 0.80 & 0.87 & 0.90 \\
\bottomrule
\end{tabular}
}
\caption{\textbf{[Correspondence-free point cloud registration]} mAA results with $5^\circ$, $10^\circ$, $15^\circ$ and $20^\circ$ thresholds for \eqref{eq:mAA} ($\uparrow$ is better. $\text{Upper bound}=1$). \underline{Underline}: Ideal accuracy obtainable if one chooses the model closest to the ground truth. \textcolor{blue}{\textbf{Blue bold}}: Best results (excluding Ideal).}
\label{tab:corfree}
\end{table}

We also demonstrate the advantages of NONSAC for the correspondence-free point cloud registration problem \cite{teaser}, where the rigid-body transformation between two point clouds is estimated by hypothesizing all-to-all correspondences, rather than relying on explicit 3D-to-3D feature matches.
Using the Bunny dataset from the Stanford 3D scanning repository \cite{bunny}, we set up the experiment as follows:
First, we scale the original point cloud data to fit inside a unit cube.
Next, we randomly select two subsets of 500 points each, with a 50\% overlap.
One subset is transformed by a random rigid body transformation, and Gaussian noise of $\sigma=0.01$ is added.
We then hypothesize all-to-all correspondences, resulting in 250,000 tentative matches with a 99.9\% outlier ratio (since the inlier ratio is $\frac{500\times0.5}{500\times500}=0.1\%$).
The size of the non-minimal sample for NONSAC is set to 10,000 points.
For the estimator, we use another modified version of PCR-99 \cite{pcr99}, which differs slightly from the variant used in Sec. \ref{subsec:pcr} and is better suited to handling a 99.9\% outlier ratio, albeit at the cost of increased computational time.
This version differs from the original \cite{pcr99} as follows:
\begin{itemize}
    \item As in Sec. \ref{subsec:pcr}, we use the pairwise distance between 3D points to prescreen bad 3-point samples and to score the correspondences.
    \item We terminate the iterations once the inlier ratio reaches at least 
0.09\% and a minimum of 1000 three-point samples have passed the prescreening test. (\ie, $1000\leq n_\mathrm{hypothesis}$ in Alg. 1 of \cite{pcr99}).
\end{itemize}

For this problem, we evaluate not only our NONSAC variants, but also several state-of-the-art point cloud registration methods, including graph-based approaches (SC2-PCR \cite{sc2pcr}, TEASER$++$ \cite{teaser}, MAC \cite{mac}, original PCR-99 \cite{pcr99}), as well as deep learning-based methods (PointDSC \cite{pointdsc} and VBReg \cite{vbreg}).

\begin{figure*}[t]
 \centering
 \includegraphics[width=\textwidth]{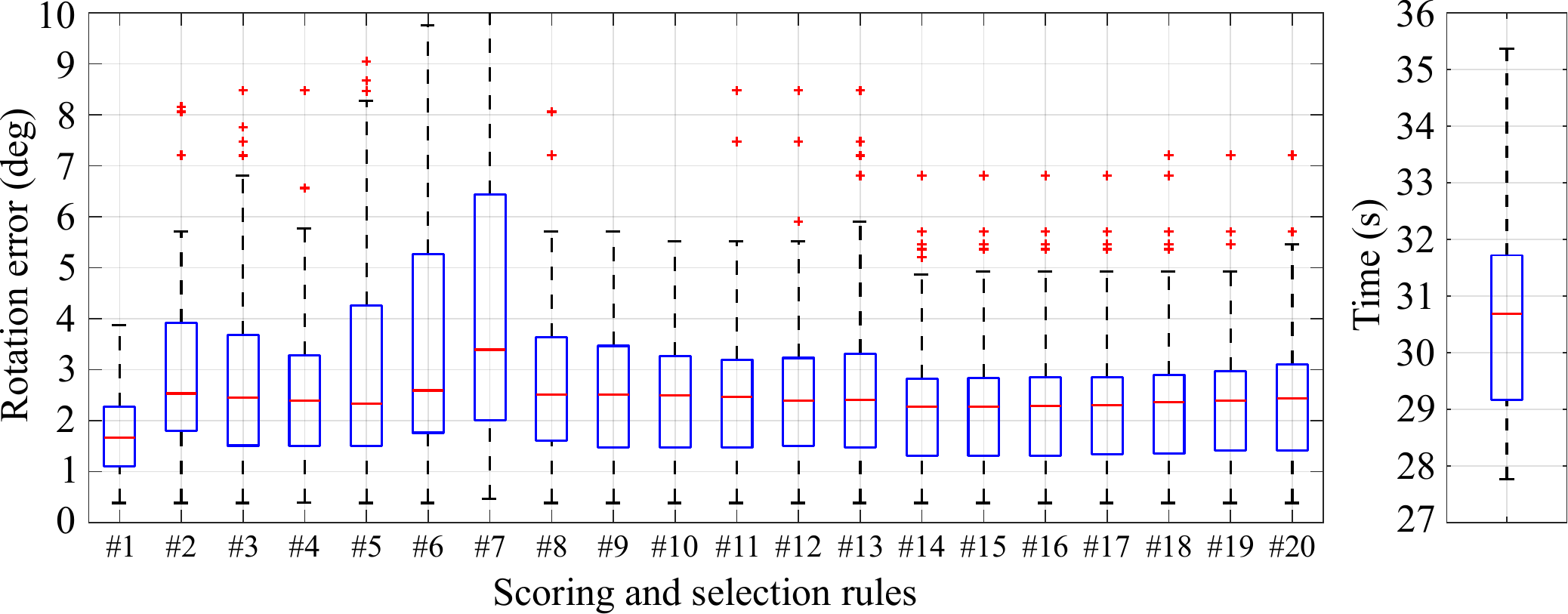}
\caption{\textbf{[Correspondence-free point cloud registration]} Left: Rotation errors of 100 Monte Carlo simulations using different scoring and selection rules for NONSAC. See Tab.~\ref{tab:corfree} for the list of rules corresponding to each ID. 
Right: Processing times for computing 10 model hypotheses.  
The experiments are conducted in MATLAB on a laptop with i7-7700HQ (2.8 GHz) and 16 GB RAM.
}
\label{fig:boxplot}
\end{figure*}

Table \ref{tab:corfree} presents the results.
Due to the sheer number of all-to-all correspondences, none of the existing methods we evaluate are able to process the full dataset. 
In contrast, NONSAC does not encounter memory limitations, as it processes the data incrementally using much smaller samples.
Regarding the scoring rule, the \textit{TLP cost ($p=0.01, 0.1$)} achieves the best performance.
As shown in Fig.~\ref{fig:boxplot}, it recovers the rigid body transformation in approximately 30 seconds, with rotation errors below $3^\circ$ in more than 75\% of the trials.

\section{Discussions}
\label{sec:discussion}

\noindent$\bullet$ \textit{Which scoring rule should be used in practice?}

In the previous section, we observed that the \textit{TLP cost ($p=0.1$--$1$)} consistently delivers the strongest overall performance. 
However, the optimal choice of $p$ depends on the specific problem and on the number of non-minimal samples used. 
If a single value must be selected without further tuning, $p=0.1$ provides a sensible starting point.

\noindent$\bullet$ \textit{Is it possible that another scoring rule could perform better?}

The performance gap between the \textit{Ideal hypothesis selection} baseline and our current results indicates that there is still room for improvement.
This may involve dynamically switching between scoring rules or adapting their parameters based on the observed residuals. 
It could also require incorporating entirely different criteria, such as measures of model uncertainty or characteristics of the inlier distribution. We leave the investigation of these directions for future work.

\noindent$\bullet$ \textit{How does one determine the appropriate size for the non-minimal sample ($N$)?}

If $N$ is too small, each non-minimal sample may contain too few inliers, leading to inaccurate model hypotheses. 
Conversely, if $N$ is too large, the variability of the inlier ratio across samples decreases, since the standard deviation of the sample proportion is approximately proportional to $1/\sqrt{N}$ (see Sec.~\ref{subsec:sampling}).
As a result, the robustness gain obtained through non-minimal sampling becomes less pronounced. 
These observations suggest that there exists a balance, or “sweet spot,” for choosing $N$.
For starters, the rule of thumb is that $N\geq I_\text{min}/P_\text{min}$, where $I_\text{min}$ and $P_\text{min}$ are respectively the minimum number of inliers and the minimum inlier ratio that the core estimator can reliably handle. 
Investigating an optimal strategy to choose $N$ is left for future work.

\noindent$\bullet$ \textit{Would increasing the number of non-minimal samples improve robustness?}

In general, robustness increases with the number of samples, as it increases the probability that at least one sample contains a sufficiently high inlier ratio to produce an accurate model hypothesis. 
However, the benefits tend to diminish beyond a certain point, as robustness becomes increasingly governed by the inherent robustness of the core estimator rather than by the number of non-minimal samples.
Moreover, computational cost grows linearly with the number of samples.
In practice, there is a trade-off between robustness and runtime, and a moderate number of non-minimal samples often provides a good balance.

\noindent$\bullet$ \textit{In which situations is NONSAC not recommended?}

NONSAC is not particularly advantageous when the dataset can be processed efficiently in a single batch, since its main strength lies in handling prohibitively large datasets. 
It may also offer limited benefit when the inlier ratio is already high, as standard robust estimators can perform well using a single non-minimal sample.
Furthermore, if the core estimator is weak, increasing the number of non-minimal samples will not compensate for its limitations. 
Finally, in time-critical applications with strict runtime constraints, the additional computational cost may outweigh the robustness gains.

\section{Conclusion}
\label{sec:conclusion}
We proposed NONSAC, a novel framework for geometric model estimation from prohibitively large datasets with high outlier ratios.
Instead of processing the full dataset at once, NONSAC repeatedly draws reasonably sized non-minimal subsets, estimates models independently on each of them, and then selects the most reliable hypothesis using a scoring rule.
This allows us to avoid the memory and computational limitations that often hinder existing methods, while also exploiting the statistical variability in inlier ratios across samples to improve robustness.
Experiments across relative pose estimation, Perspective-n-Point and point cloud registration demonstrate that NONSAC handles high outlier ratios effectively, with the truncated $L_p$ (TLP) cost offering consistently strong performance.
Overall, NONSAC offers a simple and practical paradigm for scaling robust geometric estimation to data sizes and outlier ratios that would otherwise be difficult or even impossible to handle.

\begin{table*}[ht!]
\resizebox{\textwidth}{!}{
\renewcommand{\arraystretch}{1.15}
\setlength{\tabcolsep}{3pt}
\begin{tabular}{cl*{6}{c}|*{6}{c}|*{6}{c}}
\hline
\multicolumn{2}{c}{} &
\multicolumn{6}{c}{\rule{0pt}{9pt}\# of non-minimal samples = 10} &
\multicolumn{6}{c}{\# of non-minimal samples = 20} &
\multicolumn{6}{c}{\# of non-minimal samples = 30} \\
\multicolumn{2}{c}{} &
\multicolumn{6}{c}{Outlier ratio} &
\multicolumn{6}{c}{Outlier ratio} &
\multicolumn{6}{c}{Outlier ratio} \\
\multicolumn{2}{c}{}&
65\% & 70\% & 75\% & 80\% & 85\% & 90\% & 65\% & 70\% & 75\% & 80\% & 85\% & 90\% & 65\% & 70\% & 75\% & 80\% & 85\% & 90\%\\
\hline
\multirow{21}{*}{\rotatebox{90}{$\sigma=0.002$}} & Ideal hypothesis selection & \textbf{1.00} & \textbf{1.00} & \textbf{0.99} & \textbf{0.98} & \textbf{0.92} & \textbf{0.76} & \textbf{1.00} & \textbf{1.00} & \textbf{1.00} & \textbf{0.99} & \textbf{0.97} & \textbf{0.88} & \textbf{1.00} & \textbf{1.00} & \textbf{1.00} & \textbf{1.00} & \textbf{0.98} & \textbf{0.93} \\
 & Fixed non-minimal sample & 0.99 & 0.98 & 0.97 & \cellcolor{green!30}0.96 & \cellcolor{green!30}0.87 & \cellcolor{green!30}0.65 & 0.99 & \cellcolor{green!30}\textbf{1.00} & 0.98 & \cellcolor{green!30}0.98 & 0.91 & 0.75 & 0.99 & 0.99 & 0.99 & \cellcolor{green!30}0.99 & \cellcolor{green!30}0.96 & 0.84 \\
 & Most inliers & 0.99 & 0.98 & 0.97 & 0.95 & 0.85 & 0.62 & 0.98 & 0.98 & 0.97 & 0.96 & \cellcolor{green!30}0.94 & 0.78 & 0.99 & 0.99 & 0.98 & 0.98 & 0.95 & \cellcolor{green!30}0.86\\
 & Closest pair & 0.99 & 0.98 & 0.95 & \cellcolor{red!30}0.88& \cellcolor{red!30}0.69 & \cellcolor{red!30}0.54 & 0.99 & 0.98 & 0.95 & \cellcolor{red!30}0.89 & \cellcolor{red!30}0.70 & \cellcolor{red!30}0.57 & \cellcolor{green!30}\textbf{1.00} & 0.98 & 0.97 & \cellcolor{red!30}0.89 & \cellcolor{red!30}0.75 & \cellcolor{red!30}0.69 \\
\hline
\end{tabular}
}
\caption{\textbf{[Relative pose estimation]}
mAA results ($\uparrow$ is better. $\text{Upper bound}=1$). \textbf{Bold}: Ideal accuracy obtainable if one chooses the model closest to the ground truth. \colorbox{green!30}{Green}: Best results (excluding Ideal), \colorbox{red!30}{Red}: Low accuracy (\ie, $\text{mAA}<\colorbox{green!30}{Green}-0.05$).}
\label{tab:relative}
\end{table*}
\begin{table*}[ht!]
\resizebox{\textwidth}{!}{
\renewcommand{\arraystretch}{1.15}
\setlength{\tabcolsep}{3pt}
\begin{tabular}{cl*{6}{c}|*{6}{c}|*{6}{c}}
\hline
\multicolumn{2}{c}{} &
\multicolumn{6}{c}{\rule{0pt}{9pt}\# of non-minimal samples = 10} &
\multicolumn{6}{c}{\# of non-minimal samples = 20} &
\multicolumn{6}{c}{\# of non-minimal samples = 30} \\
\multicolumn{2}{c}{} &
\multicolumn{6}{c}{Outlier ratio} &
\multicolumn{6}{c}{Outlier ratio} &
\multicolumn{6}{c}{Outlier ratio} \\
\multicolumn{2}{c}{}&
92\% & 93\% & 94\% & 95\% & 96\% & 97\% & 92\% & 93\% & 94\% & 95\% & 96\% & 97\%  & 92\% & 93\% & 94\% & 95\% & 96\% & 97\%\\
\hline
\multirow{21}{*}{\rotatebox{90}{$\sigma=0.005$}} & Ideal hypothesis selection & \textbf{1.00} & \textbf{0.98} & \textbf{0.97} & \textbf{0.93} & \textbf{0.84} & \textbf{0.72} & \textbf{1.00} & \textbf{1.00} & \textbf{1.00} & \textbf{0.99}& \textbf{0.94} & \textbf{0.84} & \textbf{1.00} & \textbf{1.00} & \textbf{1.00} & \textbf{0.99} & \textbf{0.98} & \textbf{0.92} \\
 & Fixed non-minimal sample & 0.95 & \cellcolor{green!30}0.98 & 0.96 & 0.89 & 0.74 & \cellcolor{green!30}0.69 & \cellcolor{green!30}\textbf{1.00} & \cellcolor{green!30}\textbf{1.00} & 0.99 & 0.97 & 0.91 & \cellcolor{red!30}0.66 & \cellcolor{green!30}\textbf{1.00} & \cellcolor{green!30}\textbf{1.00} & 0.99 & \cellcolor{green!30}\textbf{0.99} & 0.93 & \cellcolor{green!30}0.88 \\
 & Most inliers & \cellcolor{green!30}\textbf{1.00} & 0.97 & 0.96 & 0.90 & \cellcolor{green!30}0.82 & 0.64 & \cellcolor{green!30}\textbf{1.00} & 0.99 & 0.99 & \cellcolor{green!30}0.98 & 0.93 & 0.78 & \cellcolor{green!30}\textbf{1.00} & \cellcolor{green!30}\textbf{1.00} & \cellcolor{green!30}\textbf{1.00} & 0.98 & 0.95 & 0.86 \\
 & Closest pair & 0.97 & 0.94 & 0.93 & 0.86 & 0.77 & \cellcolor{red!30}0.51 & 0.98 & 0.98 & 0.97 & 0.95 & 0.91 & \cellcolor{red!30}0.74 & 0.99 & 0.99 & 0.98 & 0.96 & \cellcolor{red!30}0.91 & 0.85 \\
\hline
\end{tabular}
}
\caption{\textbf{[Perspective-n-Point]}
mAA results ($\uparrow$ is better. $\text{Upper bound}=1$). \textbf{Bold}: Ideal accuracy obtainable if one chooses the model closest to the ground truth. \colorbox{green!30}{Green}: Best results (excluding Ideal), \colorbox{red!30}{Red}: Low accuracy (\ie, $\text{mAA}<\colorbox{green!30}{Green}-0.05$).}
\label{tab:pnp}
\end{table*}
\begin{table*}[ht!]
\resizebox{\textwidth}{!}{
\renewcommand{\arraystretch}{1.15}
\setlength{\tabcolsep}{3pt}
\begin{tabular}{cl*{6}{c}|*{6}{c}|*{6}{c}}
\hline
\multicolumn{2}{c}{} &
\multicolumn{6}{c}{\rule{0pt}{9pt}\# of non-minimal samples = 10} &
\multicolumn{6}{c}{\# of non-minimal samples = 20} &
\multicolumn{6}{c}{\# of non-minimal samples = 30} \\
\multicolumn{2}{c}{} &
\multicolumn{6}{c}{Outlier ratio} &
\multicolumn{6}{c}{Outlier ratio} &
\multicolumn{6}{c}{Outlier ratio} \\
\multicolumn{2}{c}{}&
99.0 & 99.1 & 99.2 & 99.3 & 99.4 & 99.5 & 99.0 & 99.1 & 99.2 & 99.3 & 99.4 & 99.5 & 99.0 & 99.1 & 99.2 & 99.3 & 99.4 & 99.5\\
\hline
\multirow{21}{*}{\rotatebox{90}{$\sigma=0.01$}} & Ideal hypothesis selection & \textbf{1.00} & \textbf{1.00} & \textbf{1.00} & \textbf{1.00} & \textbf{0.97} & \textbf{0.83} & \textbf{1.00} & \textbf{1.00} & \textbf{1.00} & \textbf{1.00} & \textbf{1.00} & \textbf{0.97} & \textbf{1.00} & \textbf{1.00} & \textbf{1.00} & \textbf{1.00} & \textbf{1.00} & \textbf{1.00} \\
 & Fixed non-minimal sample & \cellcolor{red!30}0.88 & \cellcolor{red!30}0.85 & \cellcolor{red!30}0.74 & \cellcolor{red!30}0.52 & \cellcolor{red!30}0.47 & \cellcolor{red!30}0.25 & \cellcolor{red!30}0.90 & \cellcolor{red!30}0.82 & \cellcolor{red!30}0.72 & \cellcolor{red!30}0.72 & \cellcolor{red!30}0.55 & \cellcolor{red!30}0.34 & \cellcolor{red!30}0.93 & \cellcolor{red!30}0.94 & \cellcolor{red!30}0.86 & \cellcolor{red!30}0.69 & \cellcolor{red!30}0.56 & \cellcolor{red!30}0.41 \\
 & Most inliers & \cellcolor{green!30}\textbf{1.00} & \cellcolor{green!30}\textbf{1.00} & 0.99 & \cellcolor{green!30}0.99 & \cellcolor{green!30}0.96 & 0.81 & \cellcolor{green!30}\textbf{1.00} & \cellcolor{green!30}\textbf{1.00} & 0.99 & 0.99 & \cellcolor{green!30}0.99 & 0.95 & \cellcolor{green!30}\textbf{1.00} & \cellcolor{green!30}\textbf{1.00} & \cellcolor{green!30}\textbf{1.00} & 0.99 & 0.99 & 0.98 \\
\hline
\end{tabular}
}
\caption{\textbf{[Point cloud registration]}
mAA results ($\uparrow$ is better. $\text{Upper bound}=1$). \textbf{Bold}: Ideal accuracy obtainable if one chooses the model closest to the ground truth. \colorbox{green!30}{Green}: Best results (excluding Ideal), \colorbox{red!30}{Red}: Low accuracy (\ie, $\text{mAA}<\colorbox{green!30}{Green}-0.05$).}
\label{tab:pcr}
\end{table*}

\clearpage  


%
%


\begin{thebibliography}{10}
\providecommand{\url}[1]{\texttt{#1}}
\providecommand{\urlprefix}{URL }
\providecommand{\doi}[1]{https://doi.org/#1}

\bibitem{pointdsc}
Bai, X., Luo, Z., Zhou, L., Chen, H., Li, L., Hu, Z., Fu, H., Tai, C.L.: Pointdsc: Robust point cloud registration using deep spatial consistency. In: IEEE Conf. Comput. Vis. Pattern Recog. pp. 15859--15869 (2021)

\bibitem{gcransac}
Barath, D., Matas, J.: Graph-cut ransac: Local optimization on spatially coherent structures. IEEE Trans. Pattern Anal. Mach.  \textbf{44}(9),  4961--4974 (2022)

\bibitem{magsac_plus}
Baráth, D., Noskova, J., Ivashechkin, M., Matas, J.: {MAGSAC++}, a fast, reliable and accurate robust estimator. In: IEEE Conf. Comput. Vis. Pattern Recog. pp. 1301--1309 (2020)

\bibitem{briales_2018_cpvr}
Briales, J., Kneip, L., Gonzalez-Jimenez, J.: A certifiably globally optimal solution to the non-minimal relative pose problem. In: IEEE Conf. Comput. Vis. Pattern Recog. pp. 145--154 (2018)

\bibitem{Cai2024LinearRP}
Cai, Q., Li, X., Wu, Y.: Linear relative pose estimation founded on pose-only imaging geometry. arXiv  \textbf{abs/2401.13357} (2024)

\bibitem{campbell_2020_eccv}
Campbell, D., Liu, L., Gould, S.: Solving the blind perspective-n-point problem end-to-end with robust differentiable geometric optimization. In: Eur. Conf. Comput. Vis. pp. 244--261 (2020)

\bibitem{chen_2025_tpami}
Chen, H., Tian, W., Wang, P., Wang, F., Xiong, L., Li, H.: {EPro-PnP}: Generalized end-to-end probabilistic perspective-n-points for monocular object pose estimation. IEEE Trans. Pattern Anal. Mach.  \textbf{47}(11),  9413--9425 (2025)

\bibitem{trde}
Chen, W., Li, H., Nie, Q., Liu, Y.H.: Deterministic point cloud registration via novel transformation decomposition. In: IEEE Conf. Comput. Vis. Pattern Recog. pp. 6338--6346 (2022)

\bibitem{sc2pcr}
Chen, Z., Sun, K., Yang, F., Tao, W.: {SC2-PCR}: A second order spatial compatibility for efficient and robust point cloud registration. In: IEEE Conf. Comput. Vis. Pattern Recog. pp. 13211--13221 (2022)

\bibitem{loransac}
Chum, O., Matas, J., Kittler, J.: Locally optimized {RANSAC}. In: Pattern Recognition. pp. 236--243. Springer Berlin Heidelberg (2003)

\bibitem{bunny}
Curless, B., Levoy, M.: A volumetric method for building complex models from range images. In: SIGGRAPH. pp. 303--312. ACM (1996)

\bibitem{devore_statistics}
Devore, J.L., Berk, K.N., Carlton, M.A.: Modern Mathematical Statistics with Applications. Springer International Publishing (2021)

\bibitem{enqvist_2009_iccv}
Enqvist, O., Josephson, K., Kahl, F.: Optimal correspondences from pairwise constraints. In: Int. Conf. Comput. Vis. pp. 1295--1302 (2009)

\bibitem{aoxiang_2022_tpami}
Fan, A., Ma, J., Jiang, X., Ling, H.: Efficient deterministic search with robust loss functions for geometric model fitting. IEEE Trans. Pattern Anal. Mach.  \textbf{44}(11),  8212--8229 (2022)

\bibitem{fischler1981random}
Fischler, M.A., Bolles, R.C.: Random sample consensus: A paradigm for model fitting with applications to image analysis and automated cartography. Communications of the ACM  \textbf{24}(6),  381--395 (1981)

\bibitem{fredriksson_2016_cvpr}
Fredriksson, J., Larsson, V., Olsson, C., Kahl, F.: Optimal relative pose with unknown correspondences. In: IEEE Conf. Comput. Vis. Pattern Recog. pp. 1728--1736 (2016)

\bibitem{garcia_2021_jmiv}
Garcia-Salguero, M., Gonzalez-Jimenez, J.: Fast and robust certifiable estimation of the relative pose between two calibrated cameras. J. Math. Imaging Vis.  \textbf{63}(8),  1036–1056 (2021)

\bibitem{hartley_l1}
Hartley, R., Aftab, K., Trumpf, J.: L1 rotation averaging using the weiszfeld algorithm. In: IEEE Conf. Comput. Vis. Pattern Recog. pp. 3041--3048 (2011)

\bibitem{hartley_textbook}
Hartley, R., Zisserman, A.: Multiple View Geometry in Computer Vision. Cambridge University Press, 2 edn. (2003)

\bibitem{tear}
Huang, T., Peng, L., Vidal, R., Liu, Y.H.: Scalable 3d registration via truncated entry-wise absolute residuals. In: IEEE Conf. Comput. Vis. Pattern Recog. pp. 27477--27487 (2024)

\bibitem{vbreg}
Jiang, H., Dang, Z., Wei, Z., Xie, J., Yang, J., Salzmann, M.: Robust outlier rejection for 3d registration with variational bayes. In: IEEE Conf. Comput. Vis. Pattern Recog. pp. 1148--1157 (2023)

\bibitem{jin_2021_ijcv}
Jin, Y., Mishkin, D., Mishchuk, A., Matas, J., Fua, P., Yi, K.M., Trulls, E.: Image matching across wide baselines: From paper to practice. Int. J. Comput. Vis.  \textbf{129}(2),  517--547 (2021)

\bibitem{kukelova_2008_bmvc}
Kukelova, Z., Bujnak, M., Pajdla, T.: Polynomial eigenvalue solutions to the 5-pt and 6-pt relative pose problems. In: Brit. Mach. Vis. Conf. pp. 56.1--56.10 (2008)

\bibitem{lee_rsa_revisited}
Lee, S.H., Civera, J.: Robust single rotation averaging revisited. In: Eur. Conf. Comput. Vis. Worksh. pp. 30--42 (2025)

\bibitem{pcr99}
Lee, S.H., Civera, J., Vandewalle, P.: {PCR-99}: A practical method for point cloud registration with 99 percent outliers. In: Eur. Conf. Comput. Vis. Worksh. pp. 14--29 (2025)

\bibitem{lee_p3p}
Lee, S.H., Vandewalle, P., Civera, J.: {P3P} made easy. CoRR  \textbf{abs/2508.01312} (2025)

\bibitem{sift_flow}
Liu, C., Yuen, J., Torralba, A.: {SIFT Flow}: Dense correspondence across scenes and its applications. IEEE Trans. Pattern Anal. Mach.  \textbf{33}(5),  978--994 (2011)

\bibitem{long_ral_2025}
Long, C., Hu, Q., Jiang, C., Li, D., Ouyang, Z.: {BnB}-based robust pnp pose estimation method for outliers. IEEE Robotics and Automation Letters  \textbf{10}(7),  6856--6863 (2025)

\bibitem{luong_1996_ijcv}
Luong, Q., Faugeras, O.D.: The fundamental matrix: Theory, algorithms, and stability analysis. Int. J. Comput. Vis.  \textbf{17}(1),  43--75 (1996)

\bibitem{moran2024consensus}
Moran, D., Margalit, Y., Trostianetsky, G., Khatib, F., Galun, M., Basri, R.: Consensus learning with deep sets for essential matrix estimation. Advances in Neural Information Processing Systems  \textbf{37},  101529--101551 (2024)

\bibitem{bnb}
Olsson, C., Kahl, F., Oskarsson, M.: Branch-and-bound methods for euclidean registration problems. IEEE Transactions on Pattern Analysis and Machine Intelligence  \textbf{31}(5),  783--794 (2009)

\bibitem{gore}
Parra~Bustos, A., Chin, T.J.: Guaranteed outlier removal for point cloud registration with correspondences. IEEE Trans. Pattern Anal. Mach.  \textbf{40}(12),  2868--2882 (2018)

\bibitem{irls}
Peng, L., K\"ummerle, C., Vidal, R.: On the convergence of {IRLS} and its variants in outlier-robust estimation. In: Proceedings of the IEEE/CVF Conference on Computer Vision and Pattern Recognition (CVPR). pp. 17808--17818 (June 2023)

\bibitem{usac}
Raguram, R., Chum, O., Pollefeys, M., Matas, J., Frahm, J.M.: {USAC}: A universal framework for random sample consensus. IEEE Trans. Pattern Anal. Mach.  \textbf{35}(8),  2022--2038 (2013)

\bibitem{rydell_2024_cvpr}
Rydell, F., Torres, A., Larsson, V.: { Revisiting Sampson Approximations for Geometric Estimation Problems }. In: IEEE Conf. Comput. Vis. Pattern Recog. pp. 4990--4998 (2024)

\bibitem{sampson_1982_cgip}
Sampson, P.D.: Fitting conic sections to “very scattered” data: An iterative refinement of the bookstein algorithm. Computer Graphics and Image Processing  \textbf{18}(1),  97--108 (1982)

\bibitem{loftr}
Sun, J., Shen, Z., Wang, Y., Bao, H., Zhou, X.: {LoFTR}: Detector-free local feature matching with transformers. In: IEEE Conf. Comput. Vis. Pattern Recog. pp. 8918--8927 (2021)

\bibitem{acepnp}
Sun, Q., Zhang, T., Zhang, G., Wang, K., Zhu, D., Li, J., Zhang, X.: Efficient solution to pnp problem based on vision geometry. IEEE Robotics and Automation Letters  \textbf{9}(4),  3100--3107 (2024)

\bibitem{raft}
Teed, Z., Deng, J.: {RAFT}: Recurrent all-pairs field transforms for optical flow. In: Eur. Conf. Comput. Vis. p. 402–419 (2020)

\bibitem{sqpnp}
Terzakis, G., Lourakis, M.: A consistently fast and globally optimal solution to the perspective-n-point problem. In: Eur. Conf. Comput. Vis. pp. 478--494 (2020)

\bibitem{tirado_2024_CVPR}
Tirado-Gar{\'\i}n, J., Civera, J.: From correspondences to pose: Non-minimal certifiably optimal relative pose without disambiguation. In: IEEE Conf. Comput. Vis. Pattern Recog. pp. 403--412 (2024)

\bibitem{ansac}
Victor~Fragoso, Christopher~Sweeney, P.S., Turk, M.: {ANSAC}: Adaptive non-minimal sample and consensus. In: Proceedings of the British Machine Vision Conference (BMVC). pp. 43.1--43.11. BMVA Press (September 2017)

\bibitem{DEPnP}
Xie, X., Zou, D.: Depth-based efficient pnp: A rapid and accurate method for camera pose estimation. IEEE Robotics and Automation Letters  \textbf{9}(11),  9287--9294 (2024)

\bibitem{xu_2025_tip}
Xu, C., Guo, T., Huang, Y., Cheng, L.: A hough voting-based 2-point ransac solution to the perspective-n-point problem. IEEE Transactions on Image Processing  \textbf{34},  6838--6851 (2025)

\bibitem{rht}
Xu, L., Oja, E., Kultanen, P.: A new curve detection method: {R}andomized {H}ough {t}ransform ({RHT}). Pattern Recognition Letters  \textbf{11}(5),  331--338 (1990)

\bibitem{gnc}
Yang, H., Antonante, P., Tzoumas, V., Carlone, L.: Graduated non-convexity for robust spatial perception: From non-minimal solvers to global outlier rejection. {IEEE} Robotics and Automation Letters ({RA-L})  \textbf{5}(2),  1127--1134 (2020)

\bibitem{teaser}
Yang, H., Shi, J., Carlone, L.: {TEASER}: Fast and certifiable point cloud registration. IEEE Transactions on Robotics  \textbf{37}(2),  314--333 (2021)

\bibitem{yi_2018_cvpr}
Yi, K.M., Trulls, E., Ono, Y., Lepetit, V., Salzmann, M., Fua, P.: Learning to find good correspondences. In: IEEE Conf. Comput. Vis. Pattern Recog. pp. 2666--2674 (2018)

\bibitem{mac}
Zhang, X., Yang, J., Zhang, S., Zhang, Y.: 3{D} registration with maximal cliques. In: IEEE Conf. Comput. Vis. Pattern Recog. pp. 17745--17754 (2023)

\bibitem{zhao}
Zhao, J.: An efficient solution to non-minimal case essential matrix estimation. IEEE Trans. Pattern Anal. Mach.  \textbf{44}(4),  1777--1792 (2022)

\end{thebibliography}
\end{document}